%% file: midl2018.tex
\documentclass[english]{article}
\usepackage[T1]{fontenc}
\usepackage[utf8]{inputenc}
\usepackage{refstyle}
\usepackage{booktabs}
\usepackage{units}
\usepackage{textcomp}
\usepackage{amstext}
\usepackage{graphicx}
\usepackage{subscript}
\usepackage[square,sort,comma,numbers]{natbib}

\makeatletter

\AtBeginDocument{\providecommand\figref[1]{\ref{fig:#1}}}
\AtBeginDocument{\providecommand\tabref[1]{\ref{tab:#1}}}
\providecommand{\tabularnewline}{\\}
\RS@ifundefined{subsecref}
  {\newref{subsec}{name = \RSsectxt}}
  {}
\RS@ifundefined{thmref}
  {\def\RSthmtxt{theorem~}\newref{thm}{name = \RSthmtxt}}
  {}
\RS@ifundefined{lemref}
  {\def\RSlemtxt{lemma~}\newref{lem}{name = \RSlemtxt}}
  {}

\usepackage[preprint]{nips_2018}

\usepackage{tikz}
\usetikzlibrary{matrix,arrows}

\makeatother

\usepackage{babel}
\begin{document}

\title{Comparison of U-net-based Convolutional Neural Networks for Liver
Segmentation in CT}

\author{Hans Meine\thanks{additional affiliation: Fraunhofer MEVIS, Bremen, Germany, \texttt{hans.meine@mevis.fraunhofer.de}}\\
Medical Image Computing Group\\
University of Bremen, Germany\\
\texttt{meine@uni-bremen.de}\\
\And Grzegorz Chlebus\\
Fraunhofer MEVIS\\
Bremen, Germany\\
\texttt{grzegorz.chlebus@mevis.fraunhofer.de} \And Mohsen Ghafoorian\\
Diagnostic Image Analysis Group\\
Radboud University, Nijmegen, Netherlands\\
\texttt{mohsen.ghafoorian@radboudumc.nl} \And Itaru Endo\\
Department of Gastroenterological Surgery\\
Yokohama City University, Japan \And Andrea Schenk\\
Fraunhofer MEVIS, Bremen, Germany\\
\texttt{andrea.schenk@mevis.fraunhofer.de}}
\maketitle
\begin{abstract}
Various approaches for liver segmentation in CT have been proposed:
Besides statistical shape models, which played a major role in this
research area, novel approaches on the basis of convolutional neural
networks have been introduced recently. Using a set of 219 liver CT
datasets with reference segmentations from liver surgery planning,
we evaluate the performance of several neural network classifiers
based on 2D and 3D U-net architectures. 

An interesting observation is that slice-wise approaches perform surprisingly
well, with mean and median Dice coefficients above 0.97, and may be
preferable over 3D approaches given current hardware and software
limitations.
\end{abstract}

\section{Introduction}

Liver segmentation has long been investigated, as it plays an important
role when planning surgeries or catheter-based interventions~\cite{ieeepulse,endo1,heimannetal09}.
Several clinical workflows require a volumetric analysis of the liver.
The size and complex shape of the organ make manual delineation very
time-consuming. On the other hand, its appearance in computed tomography
scans varies a lot, for instance, when it contains hypo- or hyperdense
tumors, necrosis, calcifications, cirrhosis, cysts, or contrast agent.
Furthermore, several adjacent structures (e.g., heart, stomach, intestines)
may have very similar appearance, causing low boundary contrast.

Proposed methods for liver segmentation range from semi-interactive~\cite{Schenk2000Efficient}
to fully automatic~\cite{kainmuller2007shape,heimann2007shape,li2015automatic,Christ2016Automatic,Dou20163D},
and one of the first \textquotedbl challenges\textquotedbl{} organized
for a fair and reproducible comparison of segmentation methods in
medical image computing was on liver segmentation~\cite{heimannetal09}.
Many successful approaches employed statistical shape models as a
representation of typical liver shapes for guiding the segmentation
in the presence of the above difficulties \cite{heimann2007shape,kainmuller2007shape,li2015automatic}.
More recently, convolutional neural networks (CNN) have received a
lot of attention due to their ability to solve segmentation problems
formulated as voxel classification tasks with sometimes impressive
performance. Fully-convolutional network architectures (FCN) became
very popular, because they allow for very efficient training and classification
on many voxels at once~\cite{Long2015Fully}. In particular, so-called
\textquotedbl U-nets\textquotedbl{} were proposed as a deep segmentation
architecture supporting efficient end-to-end training of a multi-resolution
model in 2D~\cite{Ronneberger2015UNet} and 3D~\cite{Cicek20163D,Milletari2016VNet}.
Other methods create the multi-resolution representation in the external
preprocessing pipeline, feeding parallel low- and high-resolution
streams into the neural network, which saves GPU memory and thus allows
for larger training batches~\cite{Kamnitsas2017Efficient}. For liver
segmentation, 2D U-nets were applied slice-wise, optionally combined
with 3D conditional random fields~\cite{Christ2016Automatic}. Other
authors applied a simple 3D CNN architecture to this task~\cite{Dou20163D},
with fewer layers, larger kernels, and smaller numbers of convolutional
filters. An interesting compromise using three orthogonal 2D CNNs
was proposed for cartilage segmentation in MRI~\cite{Prasoon2013Deep}.

The goal of this work was to quantitatively and qualitatively compare
the performance of different U-net-based neural network architectures.
Our evaluation includes various existing 2D and 3D variants of U-nets,
and a novel approach using an ensemble of fully convolutional slice-wise
classifiers operating on different view directions.

\section{Methods}

\paragraph{\label{par:Image-Properties}CT Datasets}

We evaluated our method on a dataset comprising 219 liver CT volumes
for surgery planning, acquired on different scanners with a resolution
of about 0.6\,mm in-plane and a reconstructed slice thickness of
0.8\,mm. Iodinated contrast agent was applied in all cases. The data
included healthy livers, organs with benign and/or malignant liver
lesion, and pre-resected cases. Liver Segmentation and further planning
was performed on the venous phase for two types of surgery: living
donor liver transplantation~\cite{endo2ldlt}, $n=52$, 24\% and
liver resection~\cite{endo1}, $n=167$, 76\%. Patient age ranged
from 10-85 years (IQR 55-72) and 133 of the patients were male, 81
female. All cases were carefully annotated using a slice-wise approach
based on live-wire, shape-based interpolation and interactive contour
correction~\cite{Schenk2000Efficient} and were reviewed by radiological
experts.

After import, the DICOM rescale parameters were applied to get 16\,bit
Hounsfield integer values for the network's input layer. Subsequently,
for models using a fixed voxel size (2\,mm or 1\,mm, depending on
the experiment), we rescaled the volumes to an isotropic voxel size
with a Lanczos window for the CT data and a nearest neighbor interpolator
for the corresponding binary mask.

\paragraph{Training vs. Test Sets}

All available cases were randomly assigned to three disjunct groups
for training, validation, and testing. The training cases were the
largest group (comprising two thirds of the cases) and were directly
used for deriving our models. Separate validation cases were used
for evaluating performance during training. The test cases were not
used for either of these, maximizing the likelihood of being able
to predict generalization performance.

\paragraph{Neural Network Architectures}

\begin{figure}[tb]
\includegraphics[width=1\linewidth]{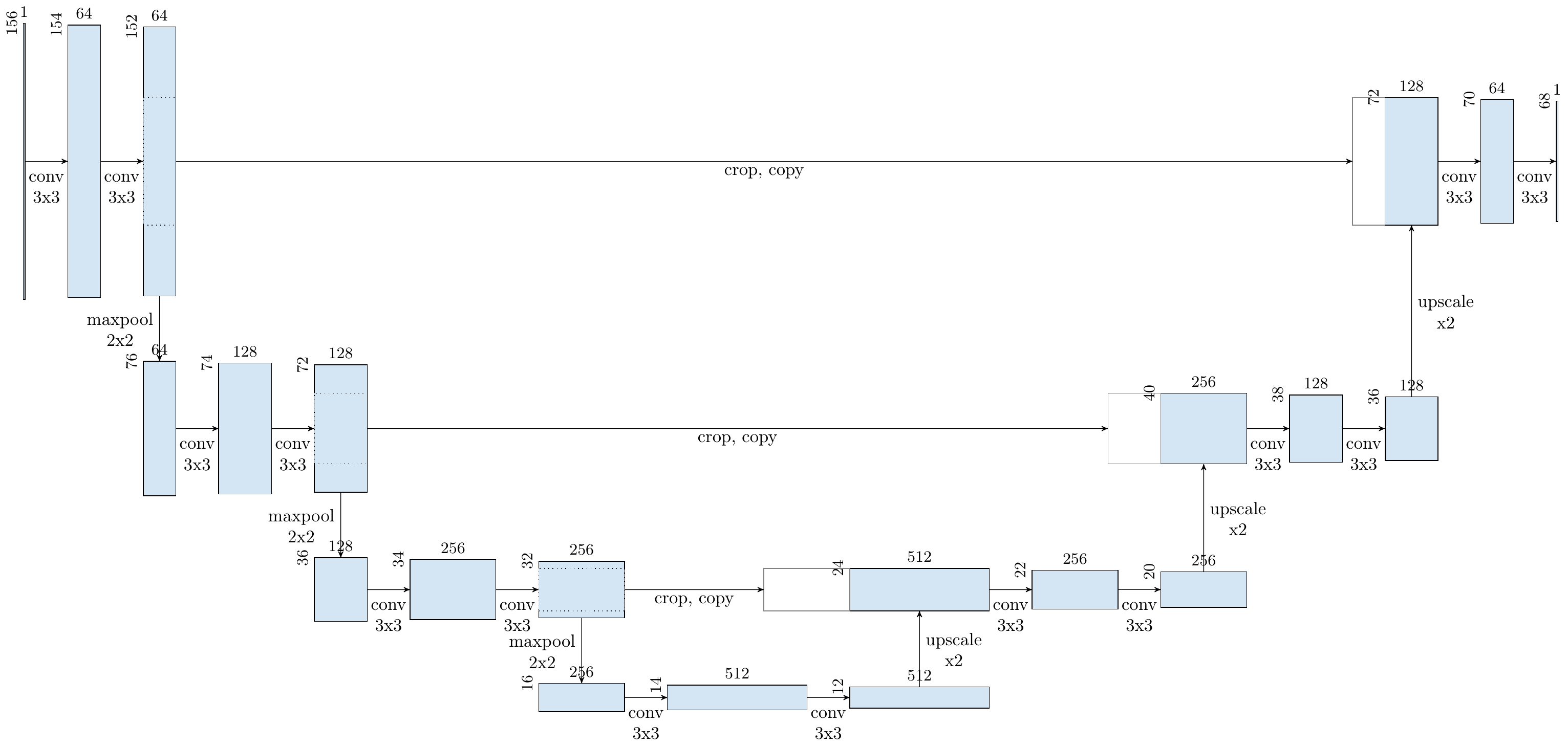}

\caption{\label{fig:U-net-architecture}Overview of U-net architecture variant
with four resolution levels (2D)}
\end{figure}
All neural network architectures discussed in this work are fully
convolutional architectures~\cite{Long2015Fully}, which means that
we can both train and apply each network efficiently on many voxels
at once by feeding larger patches into the network. The effect is
the same as training on many overlapping patches, but it is much more
efficient because many convolutions are re-used. Batch normalization
was applied before all ReLu activations for faster training.

Our baseline architecture is the \textquotedbl U-net\textquotedbl ~\cite{Ronneberger2015UNet},
named after the characteristic shape of its architecture diagram (\figref{U-net-architecture})
with a downscaling path on the left and an upscaling path on the right.
The depicted variant with four resolution levels has a receptive field
of 99\,voxels, which amounts to roughly 20\,cm after resampling
to a voxel size of 2\,mm. The size of the receptive field is important,
because it specifies the \textquotedbl window\textquotedbl{} through
which the classification is performed. For instance, if this window
is too small, the classifier cannot see the border of the liver and
performs substantially worse when the tissue appearance is anormal,
e.g. within large tumors. In order to evaluate the influence of downsampling,
we also trained networks with a reduced voxel size of 1\,mm, restoring
the U-net's fifths resolution level~\cite{Ronneberger2015UNet} to
keep the receptive field at about 20\,cm (203\,voxels). Note that
the \emph{effective} receptive field is even smaller~\cite{luo_understanding_2016},
but this is an upper bound.

Finally, we trained a 3D U-net~\cite{Cicek20163D} on this problem,
which not only increases the dimensionality of the operations, but
also requires a series of modifications: Due to memory limitations,
the number of resolution levels is only four, the number of filters
is halfed in the downscaling pathway, and the number of filters is
increased before each max pooling layer in order to prevent early
bottlenecks~\cite{Cicek20163D}. As an alternative approach to reducing
the memory requirements, we also trained a 3D U-net with zero-padding
in the convolutional layers (labeled $\text{3D}_{\text{pad}}$). This
alleviates the need for additional input padding during training and
causes the network to learn from all input voxels. During classification,
we use convolution layers \emph{without} padding, in order to prevent
boundary artifacts. We expect the benefits from larger mini-batches
to outweight the missing real image context at the patch borders.

No artificial data augmentation was performed, but the networks learned
to capture exactly the variability occurring in our data set.

\paragraph{Segmentation Approaches}

For segmentation purposes, we applied the U-nets slice-wise, without
sophisticated postprocessing (such as CRF~\cite{Christ2016Automatic}),
but discarding all connected foreground components except the largest
(in 3D). We chose the transversal viewing direction as default, but
noticed that networks trained on coronal or sagittal slices would
be able to prevent certain errors of the transversal approach. Therefore,
we also evaluated ensemble classifiers based on the softmax outputs
from three orthogonal U-nets. We report results from a simple averaging
(subsequently thresholded at $\nicefrac{1}{2}$, results labeled \textquotedbl mean\textquotedbl ),
as well as from an ensemble including a final two-layer CNN that performs
$5\times5\times5$ convolutions in 3D (128 and 64 filters, results
labeled \textquotedbl ensemble\textquotedbl ). This is similar to
previous triplanar approaches~\cite{Prasoon2013Deep}, but differs
in one important aspect: Our method is based on the softmax \emph{output}
only, allowing all four networks to be trained and executed fully
convolutionally~\cite{Long2015Fully}, which is much more efficient.

We used the Adam optimizer~\cite{Kingma2017Adam} for training the
network, and we evaluated a binary cross entropy loss function $L_{\text{CE}}$
and a Dice-based loss function $L_{\text{DSC}}$
\[
L_{\text{CE}}=-\frac{1}{N}\sum_{i}\left(y_{i}\log p_{i}+\left(1-y_{i}\right)\log\left(1-p_{i}\right)\right)\quad L_{\text{DSC}}=1-\frac{2\sum_{i}\left(y_{i}\cdot p_{i}\right)}{\sum_{i}y_{i}+\sum_{i}p_{i}}
\]
The liver typically only covers between 5~and 10\,\% of all voxels,
so the two classes are severely imbalanced. The Dice coefficient was
suggested as a novel loss function for segmentation problems~\cite{Milletari2016VNet}
because it only depends on foreground voxels (including false positives
and false negatives, but excluding all true negatives) and elegantly
alleviates the need for class balancing.

\section{Results and Discussion}

\paragraph{Quantitative Evaluation}

\begin{table}[t]
\centering\include{quantitative}

\caption{\label{tab:Median-performance-parameters}Median performance parameters
over all 40 test cases}
\end{table}
\begin{figure}[tb]
\centering\scalebox{.8}{\input{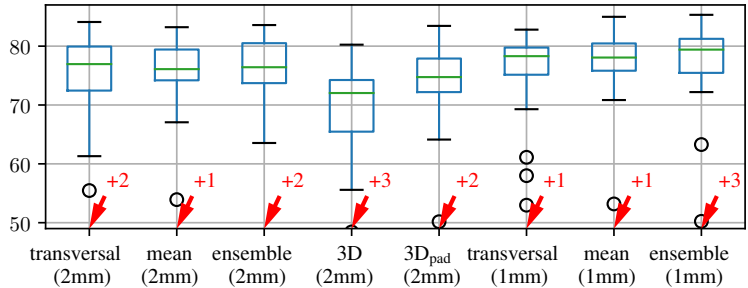}}

\caption{\label{fig:Distribution-of-MICCAI}Distribution of MICCAI scores for
all evaluated methods (red: clipped outliers)}
\end{figure}
We compared the algorithm output against the reference masks and computed
several performance criteria, the medians of which are presented in
\tabref{Median-performance-parameters}, such as the volumetric overlap
error (VOE in \%), the relative volume error $\Delta_{vol}$ in \%,
mean\,/\,max\,/\,rms surface distances in mm, and the task-specific
MICCAI score $\phi_{\text{MICCAI}}$ which averages these criteria
converted into scores in the range $\left[0\ldots100\right]$, calibrated
such that 75 is the typical performance of an untrained human observer~\cite{heimannetal09},
see \tabref{Median-performance-parameters} and \figref{Distribution-of-MICCAI}. 

\begin{figure}[tb]
\includegraphics[width=0.32\linewidth]{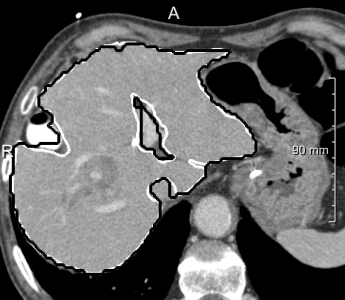}\hspace*{\fill}\includegraphics[width=0.32\linewidth]{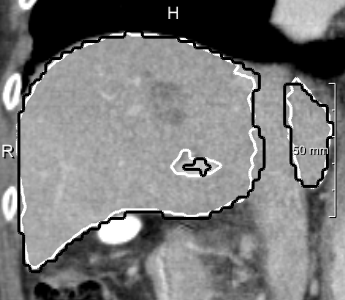}\hspace*{\fill}\includegraphics[width=0.32\linewidth]{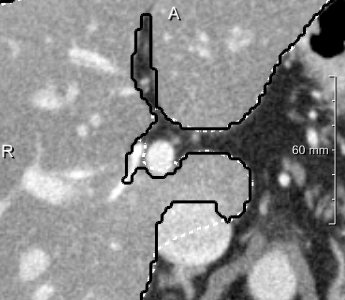}

\caption{\label{fig:ExampleCNN}Example output of U-net ensemble trained on
2\,mm voxel size (black) and reference (white); \emph{right}: application
on SLIVER07 case (dashed reference)}
\end{figure}
Example segmentations of the 2D U-net ensemble (2~mm) are illustrated
in \figref{ExampleCNN}. Contour precision is limited by the resampling,
but the model nicely excludes the vena cava and large hilar vessels
much like in our training set. This hinders comparison against the
state-of-art, since the reference masks from the SLIVER07 challenge
(dashed in \figref{ExampleCNN}, right) partially include these vessels.

\begin{figure}[tb]
\includegraphics[width=0.32\linewidth]{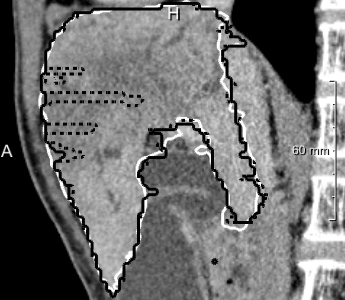}\hspace*{\fill}\includegraphics[width=0.32\linewidth]{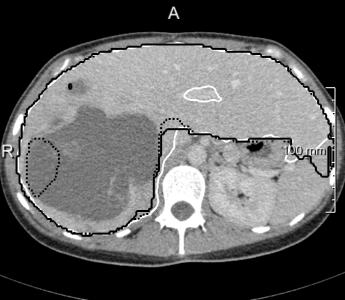}\hspace*{\fill}\includegraphics[width=0.32\linewidth]{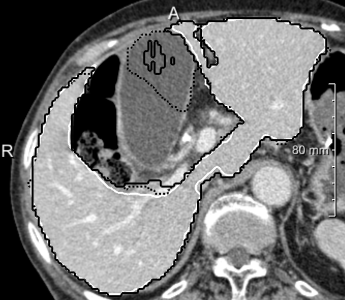}

\caption{\label{fig:ortho-vs-trans}Example problems of slice-wise approach
(dashed black) compared against ensemble of orthogonal U-nets (solid
black) and reference (white)}
\end{figure}
In most cases, the purely slice-wise application of the 2D U-net (dashed
contours in \figref{ortho-vs-trans}) does not show any comb artifacts
in orthogonal views. However, the ensemble classifier (solid contours)
performs significantly better when the appearance is severely abnormal
and 3D context is needed. In some cases, it locally performs worse,
but has an overall better volumetric overlap (Wilcoxon signed-rank
test, $p<0.05$). The ensemble models performed significantly better
than the purely 2D transversal model on the same voxel size.

\begin{figure}[tb]
\includegraphics[width=0.32\linewidth]{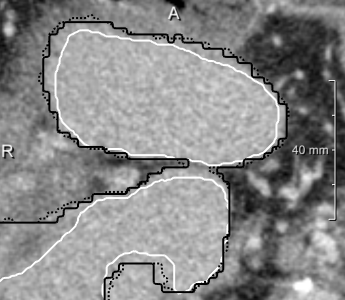}\hspace*{\fill}\includegraphics[width=0.32\linewidth]{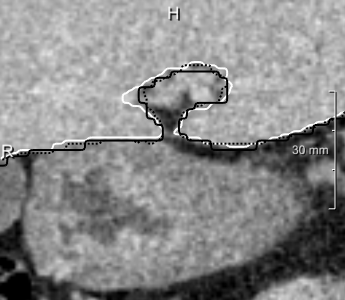}\hspace*{\fill}\includegraphics[width=0.32\linewidth]{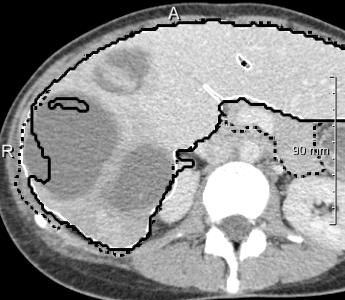}

\caption{\label{fig:examples3}\emph{Left, center}: Results from U-net ensembles
on 2\,mm (solid) and 1\,mm voxel size (dashed); \emph{right}: Comparison
of 2D U-net (solid) and 3D U-net results (dashed)}
\end{figure}

We observed some problems with narrow fissures and gaps (\figref{examples3}
left, center), and hypothesized that reducing the resampling target
voxel size from 2\,mm to 1\,mm isotropic could help here. We could
not observe qualitative changes, and the volumetric overlap did not
change significantly. The MICCAI scores, however, showed significant
improvements (Wilcoxon $p<0.05$) due to reduced surface distances,
at the expense of higher GPU memory requirements and longer training
and classification times.

\begin{table}[tb]
\centering%
\begin{tabular*}{1\linewidth}{@{\extracolsep{\fill}}lrrrrrr}
\toprule 
\multicolumn{2}{r}{mini-batch size \tikz[baseline={(0,0)}]{
  \draw[>=stealth',->]
    (0,0.35em) -| (.8em,-0.2em);
  \path[use as bounding box]
   (1em,0);
}} & max tile size & (voxels) & padded size & (voxels) & ratio\tabularnewline
\midrule
2D U-net (4 levels) & 21 & $140\times140$ & 411,600 & $228\times228$ & 1,091,664 & \textbf{38}\,\%\tabularnewline
2D U-net (5 levels) & 16 & $116\times116$ & 215,296 & $300\times300$ & 1,440,000 & 15\,\%\tabularnewline
3D U-net (4 levels)$^{\star}$ & 1 & $20\times20\times20$ & 8,000 & $108\times108\times108$ & 1,259,712 & 0.64\,\%\tabularnewline
3D U-net (4 levels) & 1 & $68\times60\times20$ & 81,600 & $156\times148\times108$ & 2,493,504 & 3.5\,\%\tabularnewline
3D U-net\textsubscript{pad} (4 levels) & 1 & $104\times104\times64$ & 692,224 & $104\times104\times64$ & 692,224 & \textbf{\emph{100}}\,\%\tabularnewline
\bottomrule
\end{tabular*}

$^{\star}$Naive variant, without reducing the number of filters in
the downscaling path~\cite{Cicek20163D}

\caption{\label{tab:Affordable-batch-sizes}Affordable batch sizes with 8 GiB
of GPU memory}
\end{table}
Lastly, we evaluated 3D U-nets, because slice-wise segmentation of
volumetric images may lead to characteristic artifacts in general.
However, both 3D U-nets performed significantly worse than the 2D
ensembles. The 3D models performed better in some of the problematic
areas of the 2D approaches (which were not many), but also brought
new problems (\figref{examples3} right). We attribute this to the
limits imposed by the available GPU memory on training batch sizes
(\tabref{Affordable-batch-sizes}). U-nets with four resolution levels
need 44 voxels of padding on each side, and when naively going from
2D to 3D (without reducing the number of filters~\cite{Ronneberger2015UNet,Cicek20163D}),
8 GiB of memory are just enough to train with mini batches containing
a \emph{single} patch of 20\textthreesuperior{} voxels each, which
does not suffice for stochastic gradient estimates stable enough for
convergence. The ratio between the number of output voxels the loss
is computed on and the number of input voxels after padding is given
in the last column of \tabref{Affordable-batch-sizes}. Consequently,
the $\text{3D}_{\text{pad}}$ U-net performed significantly better
than the 3D U-net with unpadded convolutions (Wilcoxon $p<0.05$).

With respect to the two loss functions that we investigated, we found
that the Dice loss $L_{\text{DSC}}$ did not improve performance significantly
in our experiments over binary cross entropy, even without any kind
of class balancing. We offer the following explanation of this observation,
considering a simple segmentation problem with perfect reference segmentations,
but 95\,\% background class. Although a classifier always assigning
the background class would achieve a low $L_{\text{CE}}$, all wrong
voxels do contribute to $L_{\text{CE}}$, and the gradient $\vec{\nabla}L_{\text{CE}}$
points into the right direction. Only when given faulty or imprecise
reference segmentations, or when the problem is ill-posed for some
other reason, the classifier cannot perform well and will start to
show a bias towards the more frequent class. We could observe this
problem with other datasets, but not with the expert liver segmentations
used here, which are of unusually high quality.

\section{Conclusion}

We have successfully implemented several CNN architectures for liver
segmentation in CT, based on high-quality reference segmentation from
surgery planning. We proposed an ensemble classifier comprising three
2D U-nets trained on orthogonal slices, which performed significantly
better than a single 2D U-net, and was even more robust in the presence
of abnormal cases such as resected or polycystic livers. Mean and
median volumetric overlap was above 95\,\%, the MICCAI scores were
higher than untrained humans, but apparently limited by the resampling
we performed for efficiency, which impaired the surface distances.

Surprisingly, ensembles of orthogonal 2D U-nets performed even better
than 3D U-nets trained on an NVIDIA GeForce GTX 1080 GPU, which we
attribute to limits imposed by the available memory (8~GiB). Therefore,
we conclude that a 2D classifier ensemble is an efficient approach
for the time being, in particular when taking into account deployment
scenarios in clinical environments without high-end hardware.

In the future, we want to improve the ensemble classifier by rescaling
back to the original resolution and taking the original CT into account.
Furthermore, we will investigate more efficient 3D CNNs~\cite{Kamnitsas2017Efficient}
and extend the classifier towards vessels and tumors.

\bibliographystyle{plos2015}
\bibliography{miccai2017}

\end{document}

%% file: quantitative.tex
\begin{tabular*}{.9\linewidth}{@{\extracolsep{\fill}}lrrrrrrr}
\toprule
{} &  VOE &  $\Delta_\text{vol}$ &  $d_\text{mean}$ &  $d_\text{max}$ &  $d_\text{rms}$ &  $\phi_\text{MICCAI}$ &  time [s] \\
\midrule
transversal (2mm) & 5.48 &                 3.68 &             1.02 &            19.5 &            1.68 &                  76.9 &      \textbf{3.69} \\
mean (2mm)        & 5.41 &                 4.18 &             1.04 &            18.5 &            1.69 &                  76.1 &      12.6 \\
ensemble (2mm)    & 5.21 &                 4.16 &            0.971 &            19.7 &            1.73 &                  76.4 &      13.3 \\
3D (2mm)          & 6.64 &                 4.07 &             1.25 &            22.4 &            2.41 &                    72 &       7.3 \\
3D pad (2mm)      & 5.93 &                 4.84 &             1.11 &            \textbf{18.2} &            1.81 &                  74.7 &      26.2 \\
transversal (1mm) & 5.48 &                 3.67 &            0.909 &            19.6 &            1.42 &                  78.3 &      88.1 \\
mean (1mm)        & 5.28 &                 3.97 &            0.877 &            18.5 &            \textbf{1.35} &                  78.1 &       134 \\
ensemble (1mm)    & \textbf{5.05} &                 \textbf{3.51} &            \textbf{0.874} &            19.1 &            1.44 &                  \textbf{79.4} &       146 \\
\bottomrule
\end{tabular*}